\documentclass[11pt]{article}
\usepackage[a4paper,margin=1in]{geometry}

\usepackage{graphicx}
\usepackage{amsmath}
\usepackage{amssymb}
\usepackage{multirow}
\usepackage{rotating}

\usepackage[breaklinks=true,colorlinks]{hyperref}

\usepackage{authblk}
\author[1,2]{Vasileios Sevetlidis\thanks{vsevetli@athenarc.gr}}
\author[2]{George Pavlidis}
\author[1]{Vasiliki Balaska}
\author[1]{Athanasios Psomoulis}
\author[1]{Spyridon Mouroutsos}
\author[1]{Antonios Gasteratos}
\affil[1]{Democritus University of Thrace, Xanthi, Greece}
\affil[2]{Athena Research Centre, Xanthi, Greece}

\begin{document}
\date{}

\title{Defect detection using weakly supervised learning}



\maketitle

\begin{abstract}
 In many real-world scenarios, obtaining large amounts of labeled data can be a daunting task. Weakly supervised learning techniques have gained significant attention in recent years as an alternative to traditional supervised learning, as they enable training models using only a limited amount of labeled data. In this paper, the performance of a weakly supervised classifier to its fully supervised counterpart is compared on the task of defect detection. Experiments are conducted on a dataset of images containing defects, and evaluate the two classifiers based on their accuracy, precision, and recall. Our results show that the weakly supervised classifier achieves comparable performance to the supervised classifier, while requiring significantly less labeled data.
\end{abstract}

\section{Introduction}
Defect detection is a critical task in many industries, from manufacturing \cite{li2021fabric} to healthcare\cite{castellani2020real}. Detecting defects early on can prevent quality issues and save costs by avoiding expensive rework or recalls \cite{psarommatis2020zero}. In Industry 4.0, defect detection is even more critical, as automated inspection systems are increasingly being used to identify defects in real-time, ensuring that only products of the highest quality are delivered to customers \cite{konstantinidis2023multi}. However, obtaining large amounts of labeled data for training accurate defect detection models can be a significant challenge in many scenarios \cite{konstantinidis2022assessment}. Manual labeling of data is time-consuming and expensive, while labeling errors can significantly affect the quality of the trained model \cite{sevetlidis2022tackling}.

In recent years, weakly supervised learning has emerged as a promising solution for training models using limited amounts of labeled data \cite{zhang2020survey}. Unlike traditional supervised learning, which requires a large amount of annotated data, weakly supervised learning enables models to learn from only a fraction of the labeled or even from unlabeled data \cite{zhou2018brief}. This approach has attracted significant attention due to its potential to reduce the reliance on manual annotation and improve the efficiency of the learning process. In particular, weakly supervised learning has shown promising results in defect detection tasks, where only a limited number of tagged samples are available.

This work explores the performance of a weakly supervised classifier compared to its fully supervised counterpart on the task of defect detection. The main objective is to evaluate the feasibility of weakly supervised learning in defect detection and compare it with the traditional supervised approach. Experiments are conducted on a real-world dataset of images containing defects and the two classifiers are evaluated based on their accuracy, precision, and recall. Additionally, the performance of the two classifiers under different amounts of labeled data is analyzed, to understand the effect of data availability on the performance of the models. Valuable insights will be provided by the findings into the effectiveness of defect detection tasks using weakly supervised learning, and the strengths and limitations of this approach in real-world scenarios will be identified.

\section{Related Work}
Weakly supervised learning refers to a set of machine learning techniques that aim to train models using a limited amount of labeled data, or even without any labeled data at all \cite{zhou2018brief}. In contrast to traditional supervised learning, which requires a large amount of labeled data for training, weakly supervised learning enables models to learn from weak supervision signals such as noisy labels, incomplete annotations, or partial information \cite{guo2018curriculumnet}.
One common weakly supervised learning technique is weakly supervised classification, which aims to classify input data into predefined categories without relying on fully labeled samples. Instead, these classifiers are trained using only partially annotated data, such as images with image-level tags, rather than object-level bounding boxes or pixel-level annotations. The classifier can then localize the object of interest within the image and generate saliency maps \cite{shimoda2016distinct}. Another popular technique is semi-supervised learning, which combines a small amount of labeled with a larger amount of unlabeled data to improve the model's performance. Semi-supervised learning techniques typically work by leveraging the inherent structure in the data to propagate the labels from the samples assigned with a label to the unlabeled ones \cite{zhu2022introduction}. In addition, there are other weakly supervised learning techniques, such as multi-instance learning \cite{herrera2016multiple}, where each training instance consists of a bag of instances, some labeled as positive, negative, or unlabeled. Co-training \cite{ma2017self}, which trains two or more classifiers using different views of the data, aims to improve the overall performance of the models. Finally, positive-unlabeled learning is a machine learning framework that aims to learn a binary classification model from data from a single class, e.g., the positive samples are the only ones with a label \cite{jaskie2022positive}.

Weakly supervised learning has emerged as a promising approach for defect detection, which aims to identify the presence and location of defects in products or materials. Several studies have investigated the effectiveness of weakly supervised learning on this topic. For example, \cite{xu2020weakly} proposed an approach to identify faults in steel plates. It uses image-level labels to train a classifier and generates saliency maps to localize the defects within the image. In their study, Niu et al. \cite{niu2019defectgan} investigated a surface segmentation method based on CycleGAN using a weakly supervised approach. The model was trained using image-level annotations and outperformed the supervised method on industrial datasets. \cite{pham2023defect} proposed a semi-supervised method to identify defective circuit boards (PCBs). It combined a small amount of labeled data with a large number of unlabeled ones to improve the model's performance. The approach achieved higher accuracy than its fully supervised counterpart. Another study, \cite{zhang2022image}, proposed a multi-instance learning approach for defect detection in welding seam images, which achieved comparable performance to a fully supervised model using only a small amount of labeled data.

Overall, these studies demonstrate the potential of weakly supervised learning for defect detection tasks and highlight the effectiveness of different these approaches which in general require a lot less annotated data than typical supervised learning methods.

\section{Proposed Methodology}

PU learning, also known as positive-unlabeled learning, is a subset of semi-supervised learning that deals with imbalanced classification problems where the negative samples are unknown or hard to obtain. The framework of PU learning is used in this work to simulate a weakly supervised learning setup. The framework is based on the assumption that the positive samples are accurately labeled, while the negative samples are partially labeled or unlabeled. Formally, assume there is a dataset of $N$ samples, denoted as $D = {(x_1, y_1), (x_2, y_2), ..., (x_N, y_N)}$, where $x_i$ is the $i^{th}$ input feature vector and $y_i$ is the corresponding label. However, the labels are incomplete, meaning that some of the samples are only labeled as positive $(y=1)$, while others are either unlabeled or negative. The goal of PU learning is to build a binary classifier $f(x)$ that can accurately predict the probability of a sample being positive, given the incomplete labels. 

\begin{figure*}[t!]
  \includegraphics[width=\textwidth]{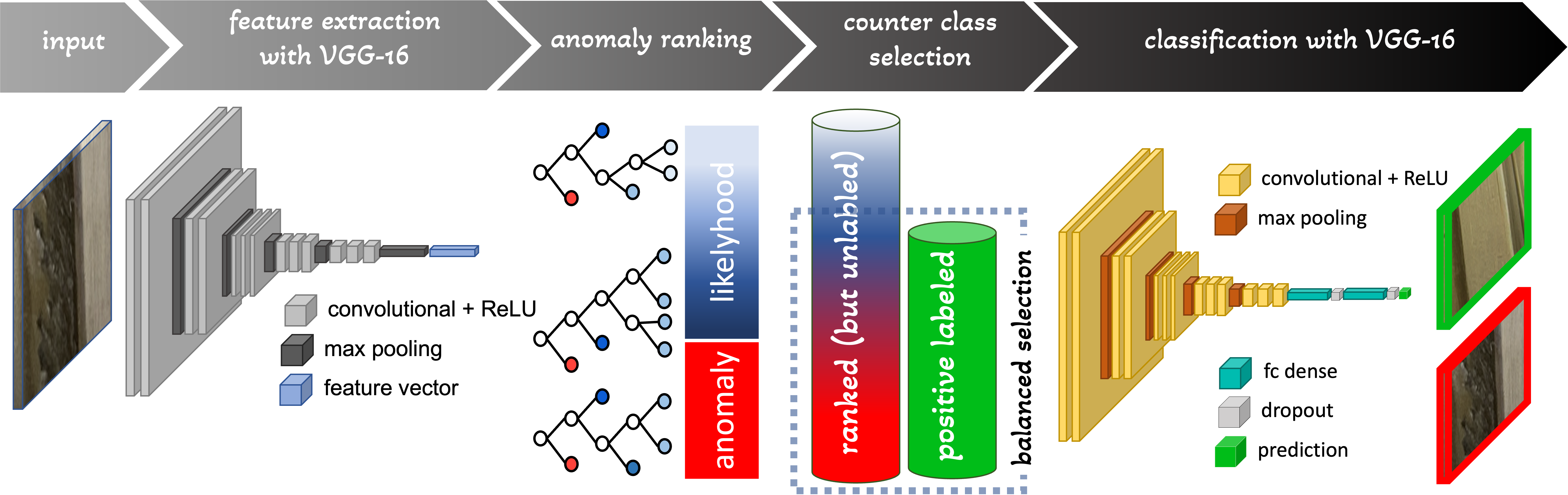}
  \caption{Overview of the proposed method}
  \label{fig:overview}
\end{figure*}

The proposed approach aims at using a small positive-labeled set with samples of a single class to assign labels to an unlabeled set which contains samples with arbitrary labels. The approach comprises a feature extractor, an anomaly scoring method, and a deep binary classifier, as shown in \figurename~\ref{fig:overview}. The feature extractor is a pre-trained deep architecture, and the anomaly scoring method is trained on the features extracted from the positive-labeled set. The resulting ranking of anomalous samples is used to create a counter-example class that is comparable to the positive-labeled set. Finally, a deep learning binary classifier is trained on the positive-labeled and counter-example sets using another deep learning model with a custom binary classifier attached.

In detail, the feature extractor is a deep learning architecture pre-trained on a large image dataset. The layers of the deep model are frozen, and no classification layers are attached to it, so its output can be used as feature vectors. In this work, a VGG-16 model is being used for this task. The images from the positive-labeled class are fed into the deep learning model, and the activation of its weights is taken as feature vectors. These feature vectors can then be used for a variety of downstream tasks, such as clustering or visualization.


The anomaly scoring method in the proposed approach is an Isolation Forest \cite{liu2008isolation}, an unsupervised machine-learning algorithm used for anomaly detection. It works by isolating anomalies in tree structures. Its decisions are based on the number of steps required to isolate them from other samples. In the framework of the proposed approach, this method is trained on the features of the positive-labeled samples and is used for scoring the feature vectors of the samples in the unlabeled set. The fewer the splits for given a sample, the less likely it is a normal instance. Thus a ranking from the most anomalous to the most normal sample is obtained.

The positive-labeled set only comprises a portion of the entire unlabeled set. If both sets were used without modification, the resulting dataset would be heavily imbalanced, which would make it unsuitable for training a supervised learning algorithm. However, a portion of the unlabeled set can be extracted that is equal in size to the positive-labeled set. By using the ranked anomaly scores, a counter-example class can be created, as the most anomalous samples are included in this set.

Finally, a deep learning binary classifier is trained on the positive-labeled and the counter-example sets. Another VGG16 architecture is used with a custom binary classifier attached to it. 

\section{Experiments}

This work involves comparing a supervised learning binary classifier that was trained using the entire dataset with different variants of the proposed weakly supervised approach. The variants test the performance of the approach using different initial populations in the positive-labeled class, ranging from 5\% to 30\% of the non-defective data depending on the experimental setup. All experiments used a 5-fold cross-validation and data split 80-20 between training and testing. It's worth noting that the supervised learning binary classifier and the binary classifier at the final step of the proposed approach share the same architecture to enable a direct comparison of their performance under the different learning frameworks.

\subsection{Description of the dataset}
 A \textit{ball screw drive} is a mechanical system that converts rotary motion into linear. It consists of a threaded shaft and a series of ball bearings that run along the threads. The ball nut contains a threaded hole that matches the thread on the ball screw and is held in place while the ball screw rotates. As the ball screw rotates, the ball bearings inside it move along the threads, pushing the nut along the shaft and creating linear motion. Ball screw drives are commonly used in applications requiring precise, repeatable linear motion, such as machine tools, robotics, and automation equipment  \cite{schlagenhauf2019camera}.
 
Pitting is a type of surface damage that can occur in ball screws. It is a form of fatigue failure characterized by the formation of tiny, localized craters or pits on the surface of the ball screw. Pitting is typically caused by repeated cyclic loading of the ball screw, which can result in the accumulation of small cracks or defects in the material. Over time, these cracks can grow and merge, leading to the formation of pits on the surface of the ball screw. In this work, the goal is to distinguish Pitting from intact components or lubrication traces.

\begin{figure*}[t!]
  \includegraphics[width=\textwidth]{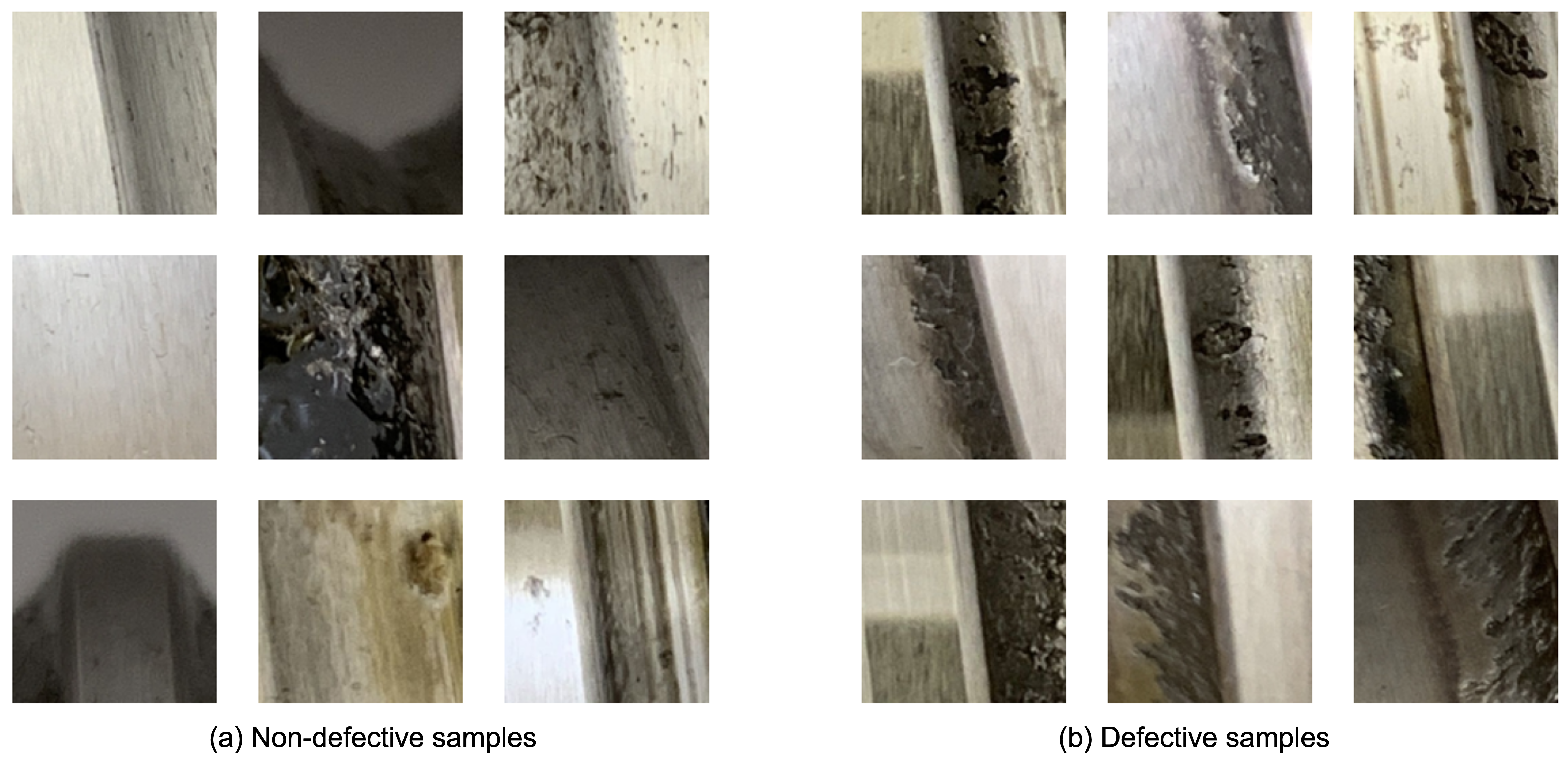}
  \caption{Random samples from the BSD dataset. Non-defective samples might exhibit lubrication signs raising a false alarm in defect detection systems.}
  \label{fig:dataset}
\end{figure*}

Specifically, the \textit{Ball Screw Defect for Classification} (BSD) \cite{schlagenhauf2021industrial} dataset is publicly available\footnote{The dataset can be found here: \url{https://publikationen.bibliothek.kit.edu/1000133819}} and consists of 21,835 RGB images with a resolution of 150x150 pixels, depicting the surface of Ball Screw Drives. Among these images, 11,075 are without surface defects, while the remaining images show surface defects in the form of pittings. This distribution ensures an equal representation of both classes in the dataset; a sample of BSD is shown in \figurename~\ref{fig:dataset}. As mentioned before, it is essential to identify surface defects promptly to maintain machine availability. This dataset allows researchers and practitioners to train and test machine learning models to classify surface defects on machine tool elements accurately.

This dataset provides labels for all samples. In order to test the proposed approach in the weakly supervised setting, it needs to be adjusted accordingly. Thus, the data are split into a positive-labeled set comprising a small percentage of samples coming from one class, for example the non-defective class, and the remaining samples along with the unused class, e.g., the defective samples, become the unlabeled ones.

\subsection{Data preprocessing}

The images were rescaled from $150\times 150$ to $128 \times 128$ and they were normalized so their values ranged in $[0,1]$. The feature vectors were used as input at the anomaly detection method without further modifications. Dataset augmentation techniques were applied to the last step (binary classification) to combat inadequate training due to the small training set sizes. The augmentations were twofold: (a) random affine transformations, including rotations, vertical and horizontal flipping, shearing, cropping, and contast adjustment; and (b) sample mixing with MixUp \cite{zhang2017mixup} with default parameters.

\subsection{Configuration}

As aforementioned, the feature extractor is a VGG16 architecture \cite{simonyan2014very} pre-trained on the ImageNet\cite{deng2009imagenet}. It was chosen because it is a prevalent deep-learning model which has been used numerous times for benchmarking since its inception. Moreover, it is easy to train and does not exhaust computational resources. The layers of the deep model are frozen, and no classification layers are attached to it. The model's output is flattened, so a feature vector is obtained $\mathbb{R}^{1\times 8192}$. 

Isolation Forest was chosen because it is fast to train due to its random splits, adapts well in highly non-linear spaces, and it is easy to optimize its hyper-parameters (i.e., the number of estimators, maximum depth, contamination ratio, and maximum sample size) \cite{liu2008isolation}. Hence, the number of estimators was set to $100$; the depth was left at the default operation, the number of samples per estimator was $256$, and the contamination fraction at $C=0.1$.

The proposed approach\footnote{The proposed method was built in Python using Keras as the deep learning backbone, and all experiments were executed on the same platform.} also utilizes a VGG16 architecture as the binary classifier. It was the same architecture used by the authors who introduced the BSD dataset and set a baseline for supervised binary classification. The additional custom classification layers were two Dense, fully connected layers with ReLU activation, a Dropout layer of $0.2$ rate between them, and a final fully connected layer with sigmoid activation. Using the same model and the proposed hyperparameters \cite{schlagenhauf2019camera} allows for a direct comparison of the performance of our supervised learning instance, the weakly supervised one, and theirs. Note that all model parameters are trainable, and they were trained from scratch.

\section{Results}



\begin{table}[t]
{\footnotesize
\centering
\setlength\tabcolsep{1pt}
\caption{A comparison between supervised learning and weakly supervised performances for the BSD dataset. Also, the performance of the proposed method is evaluated for different initial populations.}\label{table:results}
\begin{tabular}{cccclcccccc}
\hline\hline
 &  & \multicolumn{3}{c}{Supervised Learning}                &  & \multicolumn{5}{c}{Weakly Supervised Learning}                                \\  \cline{3-5}\cline{3-5} \cline{7-11}\cline{7-11} 
    Positive-labeled   &  & 100\%              & \multicolumn{2}{c}{100\%}         &  & 5\%           & 10\%          & 15\%          & 20\%          & 30\%          \\ \cline{1-1}\cline{1-1} \cline{3-5}\cline{3-5} \cline{7-11}\cline{7-11} 
Accuracy (\%)    &  & 91.5 \cite{schlagenhauf2019camera}          & \multicolumn{2}{c}{96.68 (±0.09)} &  & 80.35 (±1.56) & 88.35 (±0.96) & 90.39 (±0.81) & 93.29 (±1.05) & 93.42 (±0.91) \\
Precision (\%)   &  & 93.68 \cite{schlagenhauf2019camera}         & \multicolumn{2}{c}{96.09 (±0.15)} &  & 76.34 (±1.11) & 87.34 (±0.71) & 93.22 (±0.41) & 93.12 (±0.93) & 93.02 (±1.04) \\
Recall (\%)      &  & 89.00 \cite{schlagenhauf2019camera}         & \multicolumn{2}{c}{97.03 (±0.11)} &  & 81.65 (±1.48) & 88.29 (±0.84) & 88.51 (±0.33) & 90.86 (±0.79) & 91.23 (±0.64) \\
F1-score (\%)    &  & 91.28              & \multicolumn{2}{c}{96.47 (±0.18)} &  & 78.90 (±1.33) & 87.81 (±0.80) & 90.43 (±0.74) & 91.98 (±0.99) & 92.14 (±0.83) \\ \hline\hline
\end{tabular}
}
\end{table}

The \tablename~\ref{table:results} provides a comparison of the performance metrics (accuracy, precision, recall, and F1-score) of a supervised learning approach versus the proposed approach with different amounts of positive-labeled data. The supervised learning approach achieves the highest accuracy, precision, recall, and F1-score, but requires a large amount of labeled data. On the other hand, the weakly supervised learning approaches perform reasonably well even with a small amount of labeled data, but their performance is lower than the supervised learning approach.

The results in the table also show that increasing the percentage of positive-labeled data generally leads to better performance for the proposed approach. For instance, even with just 5\% of positive-labeled data, the proposed approach achieves an accuracy score of around 80\%. This suggests that the proposed weakly supervised learning method can be useful in scenarios where obtaining large amounts of labeled data is not feasible or expensive.

It is also worth noting that the performance of the weakly supervised learning approach generally improves as the percentage of positive-labeled data increases. For instance, the F1-score increases from 78.90\% to 92.14\% as the percentage of positive-labeled data increases from 5\% to 30\%. The latter suggests that the proposed method can benefit from more labeled data, although they still may not match the performance of supervised learning methods.

Additionally, there is some variability in the performance of the weakly supervised learning approach across different percentages of positive-labeled data. For example, the precision scores increase as the percentage of positive-labeled data increases, but the recall scores show a small decrease in some cases. Overall, the table provides useful information about the trade-off between the amount of labeled data required for supervised learning versus the performance of weakly supervised learning approach with different amounts of positive-labeled data.

\section{Conclusion and Future Work}

Pitting can have a negative impact on the performance and reliability of ball screws. It can cause increased friction and wear, reduced accuracy, and increased noise levels. In severe cases, pitting can lead to complete failure of the ball screw \cite{schlagenhauf2019camera}. 
The proposed approach of weakly supervised learning presents an effective solution for assigning labels to an unlabeled set with samples of arbitrary labels. By using a small positive-labeled set, a feature extractor, an anomaly scoring method, and a deep binary classifier, the approach is able to achieve impressive results. The feature extractor, a pre-trained deep architecture, is used to extract feature vectors, while the anomaly scoring method, Isolation Forest, is used to detect anomalies and rank samples. Finally, a deep learning binary classifier is trained on the positive-labeled and counter-example sets, resulting in accurate labeling of the unlabeled set. 

This study highlights the potential of weakly supervised learning as a valuable alternative to traditional supervised learning methods, especially in scenarios where labeled data is scarce. Further investigations can be conducted on other image-based tasks and datasets to provide valuable insights into the effectiveness and generalizability of the approach. Moreover, incorporating domain knowledge or prior information into the weakly supervised learning framework can potentially enhance the performance of the method. Finally, exploring the potential of combining weakly supervised learning with other machine learning techniques, such as transfer learning or active learning, can lead to more effective and efficient approaches for solving real-world problems with limited labeled data.

\section*{Acknowledgements}
This research has been co-financed by the European Union and Greek national funds through the Operational
Program Competitiveness, Entrepreneurship and Innovation, under the call RESEARCH - CREATE - INNOVATE
grant number T2EDK-01658.

{\small
\bibliographystyle{ieeetr}
\bibliography{arxiv}
}

\end{document}